\documentclass[sigconf,nonacm]{acmart}

\AtBeginDocument{%
  }

\setcopyright{none}                
\usepackage{tabularx}

\usepackage{multirow} 

\usepackage{subcaption}

\begin{document}

\title{RegGuard: AI-Powered Retrieval-Enhanced Assistant for Pharmaceutical Regulatory Compliance}

\author{Siyuan Yang}
\affiliation{%
  \institution{Xi'an Jiaotong-Liverpool University}
  \city{Suzhou}
  \country{China}
}
\email{Swain.Yang@outlook.com}

\author{Xihan Bian}
\authornote{Corresponding author.}
\affiliation{%
  \institution{Xi'an Jiaotong-Liverpool University}
  \city{Suzhou}
  \country{China}
}
\email{Xihan.Bian@xjtlu.edu.cn}

\author{Jiayin Tang}
\affiliation{%
  \institution{Roche Diagnostics (Suzhou)}
  \city{Suzhou}
  \country{China}
}
\email{Casper.tang@roche.com}

\begin{abstract}
The increasing frequency and complexity of regulatory updates present a significant burden for multinational pharmaceutical companies. Compliance teams must interpret evolving rules across jurisdictions, formats, and agencies—often manually, at high cost and risk of error. We introduce RegGuard, an industrial-scale AI assistant designed to automate the interpretation of heterogeneous regulatory texts and align them with internal corporate policies. The system ingests heterogeneous document sources through a secure pipeline and enhances retrieval and generation quality with two novel components: HiSACC (Hierarchical Semantic Aggregation for Contextual Chunking) semantically segments long documents into coherent units while maintaining consistency across non-contiguous sections. ReLACE (Regulatory Listwise Adaptive Cross-Encoder for Reranking), a domain-adapted cross-encoder built on the open-source model, jointly models user queries and retrieved candidates to improve ranking relevance.Evaluations in enterprise settings demonstrate that RegGuard improves answer quality specifically in terms of relevance, groundedness, and contextual focus, while significantly mitigating hallucination risk. The system architecture is built for auditability and traceability, featuring provenance tracking, access control, and incremental indexing, making it highly responsive to evolving document sources and relevant for any domain with stringent compliance demands.

\end{abstract}


\maketitle

\section{Introduction}
The pharmaceutical industry operates within a highly regulated environment due to its products directly impacting human health. Ensuring legal compliance is crucial to safeguard patient safety, maintain public trust, and provide safe and effective medications~\cite{Kher2020}. Non-compliance can lead to significant financial losses, legal penalties, and reputation damage. In 2023, the U.S. Food and Drug Administration (FDA) issued 1,150 warning letters regarding drug compliance issues~\cite{sharma2023regulatory}, and in 2024, the average cost per violation reached \$14.8 million. For pharmaceutical companies, maintaining compliance that exceeds industry standards is not just about avoiding fines; but it is essential for their survival in a highly competitive market.

Regulations continue to evolve rapidly to address advancements in biotechnology and market trends. In 2024, the FDA revised 15\% of drug manufacturing regulations to better adapt to new biotechnological advances~\cite{fda2024cfr211}. However, this rapid regulatory change has led to an increasing shortage of skilled professionals capable of managing the complexity of regulatory requirements. A recent survey showed that 56\% of pharmaceutical companies reported difficulties in hiring such talent~\cite{compliancequest2025regulatory}. This situation has led to an increased reliance on automation tools to handle intricate compliance processes, such as Document Management Systems (DMS) and compliance software, which help automate tasks like record-keeping, risk assessment, and monitoring of regulatory changes~\cite{jordan2022document}. Yet, despite the availability of these tools, experts and compliance officers still struggle to efficiently track and adapt to continuous updates across global jurisdictions and diverse industry segments.

To address these challenges, we develop RegGuard, an AI-driven interactive dialogue system powered by Large Language Models (LLMs). These models are capable of processing vast amounts of legal documents and regulatory updates, helping businesses understand and respond to changes in a shorter time frame. Specifically, the system parses regulatory texts, guidance documents, and industry standards, assisting companies in adapting to evolving regulatory requirements across different global regions. However, one critical issue limiting LLM is the tendency to hallucination biases, where the model generates plausible but incorrect information~\cite{ji2023survey}. In high-stakes industries like pharmaceuticals, even minor inaccuracies in regulatory interpretations can result in noncompliance, legal penalties, and irreversible harm to patient safety.

To mitigate the hallucination bias of LLM, we introduce a solution based on Retrieval-Augmented Generation (RAG) technology. In collaboration with Roche, we leverage their internal compliance reports, Standard Operating Procedures (SOP) documents, and quality control records to build a system that integrates two key innovations: HiSACC(Hierarchical Semantic Aggregation for Contextual Chunking), a hierarchical semantic chunking method, ReLACE (Regulatory Listwise Adaptive Cross-Encoder for Reranking), a domain-adapted cross-encoder trained with listwise ranking on in-house regulatory QA data.HiSACC optimizes the chunking process by dynamically identifying semantically meaningful segments in regulatory documents, while ReLACE enhances retrieval performance by improving post-retrieval ranking to better match query contexts. By optimizing both pre-retrieval filtering and post-retrieval validation algorithms, our system significantly improves the accuracy and timeliness of compliance checks. This not only improves operational efficiency but also reduces the risk of regulatory penalties and enhances the reliability of compliance-related decisions.

The effectiveness of our system has been systematically evaluated through a hybrid assessment process, where large language models act as primary judges complemented by compliance officers’ verification.While our current work focuses on the pharmaceutical domain, the system’s adaptable framework also holds potential for broader applications in other compliance-sensitive contexts.

\section{Related Work}
The integration of Large Language Models (LLMs) into enterprise compliance workflows introduces fundamental challenges in data integrity, traceability, and system reliability. While LLMs demonstrate strong capabilities in document understanding and synthesis, their susceptibility to hallucination—generating factually inconsistent or unsupported content—poses unacceptable risks in regulated domains~\cite{hallucination2022}. Three primary industrial strategies have emerged to mitigate this risk: prompt engineering, retrieval-augmented generation (RAG), and domain-specific fine-tuning. Prompt engineering, despite its low implementation overhead, lacks robustness across heterogeneous compliance document types and does not provide explicit data lineage guarantees~\cite{prompt2020}. Domain-specific fine-tuning can improve specialization but requires substantial computational resources and adapts poorly to rapidly evolving regulatory corpora~\cite{finetune2022}. As a result, RAG has become the dominant industrial architecture, offering a modular workflow in which real-time retrieval grounds model outputs in authoritative sources~\cite{lewis2020retrieval}. Despite these advantages, naive RAG pipelines still fall short in compliance-critical settings: regulatory documents are lengthy, heterogeneous, and frequently updated, and static chunking combined with generic embeddings often fails to preserve semantic structure and retrieval accuracy~\cite{gao2024retrieval}. Recent research therefore focuses on improving three core stages of the RAG pipeline—chunking, pre-retrieval, and post-retrieval—but these methods typically optimize isolated components and are rarely integrated into end-to-end, production-grade systems.
\subsection{Chunking Strategies}
Chunking controls the granularity at which regulatory documents are indexed and retrieved. Conventional approaches segment text using simple heuristics such as fixed-length windows, sentence boundaries, or paragraph breaks. Although efficient, these rule-based strategies often break apart semantically cohesive clauses or merge unrelated content, which degrades retrieval quality and downstream reasoning~\cite{gao2024retrieval}. To address this, several advanced chunking methods have been proposed. LumberChunker uses a pre-trained LLM to detect semantic shift points and adapt chunk boundaries accordingly, producing contextually coherent segments~\cite{duarte2024lumberchunker}. Meta-Chunking exploits margin sampling and perplexity-based signals to capture logical transitions in long-form text~\cite{zhao2024meta}. For visually rich documents, VisRAG incorporates layout and image information via a visual language model to align chunks with structural and visual cues~\cite{yu2024visrag}. These dynamic chunking strategies improve semantic segmentation and provide more faithful retrieval units.

\subsection{Pre-Retrieval Optimization}
Pre-retrieval optimization focuses on improving the alignment between user queries and the indexed document space, thereby increasing the likelihood of retrieving semantically relevant content. Prior work has explored several techniques for refining query representations before retrieval.

\paragraph{Query Rewriting} Query rewriting adjusts long-tail or underspecified queries into forms that better match index terminology, improving recall in retrieval systems~\cite{wang2024maferw}.

\paragraph{Query Expansion} Query expansion augments a query with semantically related terms—such as synonyms or hypernyms—to broaden the search space and capture a more comprehensive set of relevant documents~\cite{koo2024optimizing}.

\paragraph{Query Transformation} Query transformation restructures query syntax or semantics while preserving intent, improving compatibility with embedding models and downstream retrieval components~\cite{chan2024rq}.

\subsection{Post-Retrieval Optimization}
Post-retrieval optimization refines the retrieved candidate set to ensure that only contextually relevant information is passed to the generation stage. Existing work primarily improves this process through context compression and relevance filtering.

\paragraph{Efficient Compression}
Compression-based methods aim to condense multiple retrieved fragments into compact representations that preserve essential semantics. COCOM, for instance, uses a Transformer architecture to merge retrieved text blocks into a single dense vector, reducing redundancy and lowering sensitivity to irrelevant evidence~\cite{rau2024context}. Such approaches help control context length while retaining useful information for downstream generation.

\paragraph{Relevance Filtration}
Filtration methods selectively identify and retain the most pertinent fragments from the retrieved set. E2E-AFG employs an adaptive filtering mechanism to discard irrelevant content~\cite{e2eafg2024}, while techniques based on string inclusion, lexical overlap, or cross-mutual information estimate a fragment’s alignment with the intended answer~\cite{Es2023RAGAS}. Other work incorporates structured semantic representations—such as Abstract Meaning Representation (AMR)—to emphasize core concepts and suppress noise, particularly in specialized domains~\cite{shi2024amr}.

\section{System Architecture and Deployment}

\begin{figure}[htbp]
    \centering
    \includegraphics[width=\linewidth]{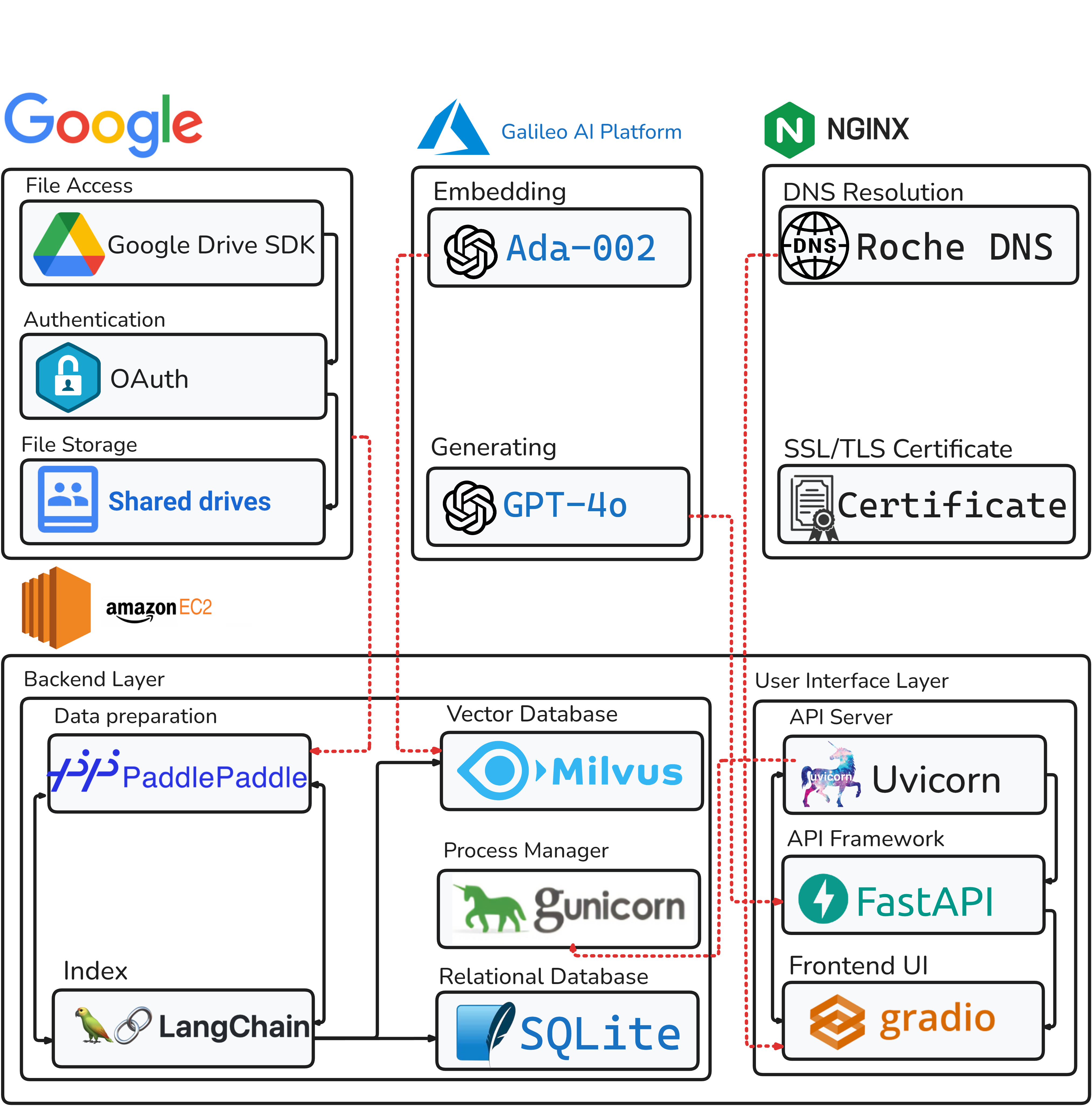}
    \captionsetup{font=small} 
    \caption{System Architecture Overview of RegGuard}
    \label{infra}
\end{figure}

RegGuard is designed as an enterprise-grade retrieval-augmented generation (RAG) system that supports large-scale regulatory compliance analysis across heterogeneous document sources. Its architecture is driven by four key industrial requirements: (1) reliable ingestion of multi-format and frequently updated documents, (2) semantically faithful retrieval over long and cross-referenced regulatory texts, (3) secure and compliant deployment within internal enterprise infrastructure, and (4) robust end-to-end operation with automatic fault recovery. Figure~\ref{infra} provides an overview of the system architecture.

\subsection{Data Acquisition and Synchronization}
Roche uses Google Workspace as its primary collaboration platform, and most compliance-relevant documents are maintained in enterprise Google Shared Drives. RegGuard ingests these documents via the Google Drive API using OAuth~2.0 authentication~\cite{googleDriveDownloads,googleDriveAboutSDK} for monitoring statistics over 30 days. A local SQLite database tracks file metadata—including timestamps and checksums—to support incremental synchronization. Each synchronization cycle detects newly added, updated, or deleted files while avoiding redundant reprocessing.


\subsection{File Conversion and Normalization}

The system supports multi-format documents, including PDF, DOC/DOCX, XLS/XLSX, CSV, TXT, as well as Google Docs and Google Sheets. Standard PDF, DOCX, and TXT files are parsed directly. For legacy DOC files and spreadsheet formats (XLS/XLSX), the system applies LibreOffice in headless mode to convert them into DOCX and CSV respectively, enabling consistent and reliable text extraction. Google Docs and Google Sheets are exported through the Google Drive API. For scanned or image-based documents, PaddleOCR is used to extract textual content prior to normalization. All documents are ultimately standardized into structured text blocks to ensure uniformity during semantic segmentation and indexing.

\subsection{Hierarchical Recursive Chunking}

Regulatory documents often contain long sections, nested structures, and dense cross-references that make simple fixed-length or delimiter-based chunking insufficient for reliable retrieval. RegGuard therefore employs a hierarchical recursive chunking strategy designed to preserve semantic continuity while meeting token constraints. The process first segments documents using structural cues—such as paragraphs and whitespace—and then applies controlled overlaps to maintain context across segment boundaries. In addition, lightweight semantic checks are used to avoid splitting closely related content, producing chunks that better reflect the logical flow of regulatory texts.

\subsection{Embedding and Indexing}

Each text segment is embedded using {\ttfamily text-embedding-ada-002} model accessed through Roche’s internal Galileo AI Platform, which wraps Azure OpenAI Service. When large batches of documents require embedding, the platform’s conservative rate limits make HTTP~429 responses more likely. To maintain throughput, the system monitors rate-limit response headers to determine appropriate wait times and falls back to an exponential backoff strategy when required. Embeddings and associated metadata are stored in a Milvus vector database for downstream retrieval.

\subsection{Retrieval and Generation Pipeline}

\begin{figure}[htbp]
    \centering
    \includegraphics[width=\linewidth]{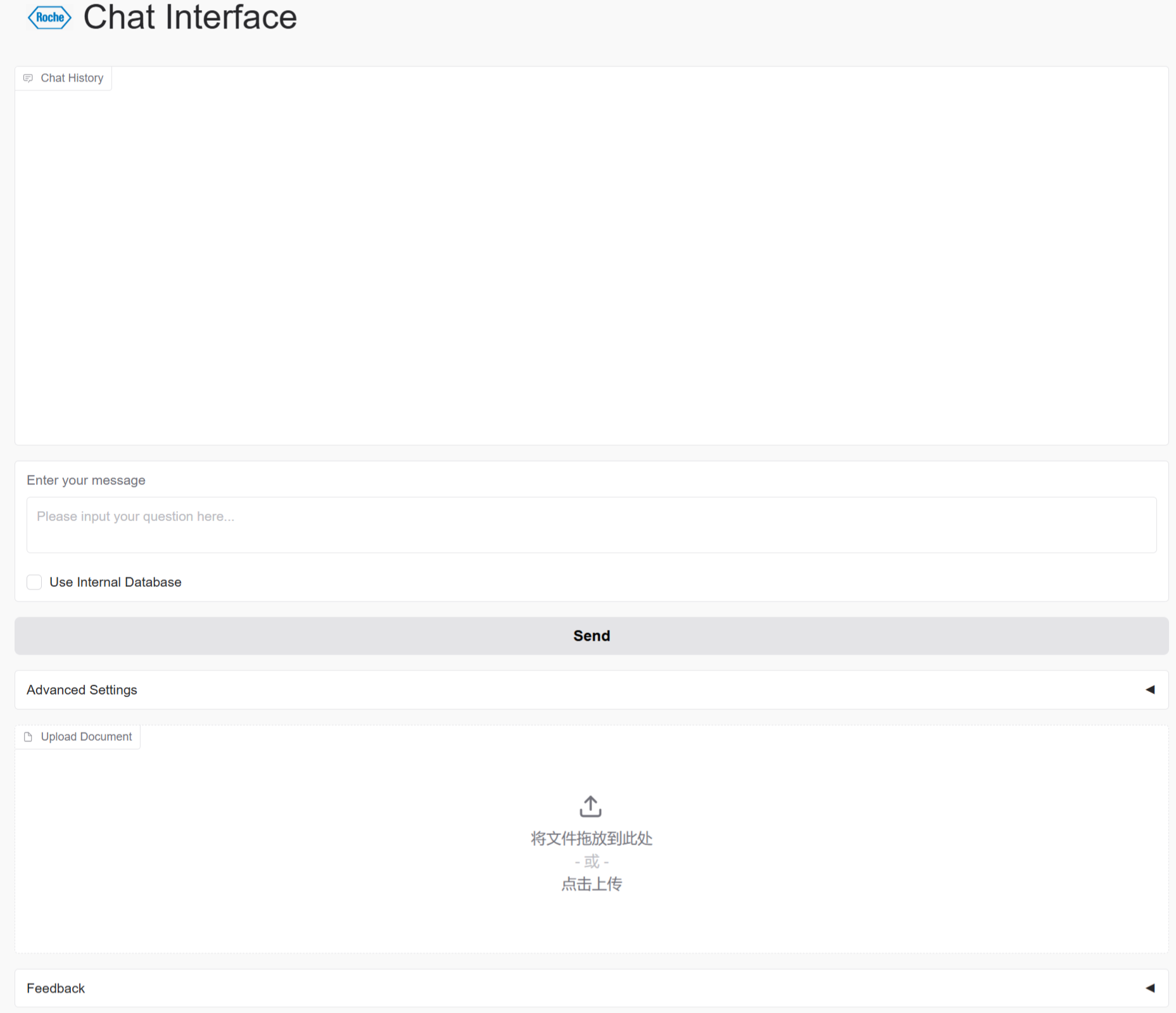}
    \captionsetup{font=small} 
    \caption{Gradio-based front-end interface}
    \label{fig:gradio-ui}
\end{figure}

Users interact with RegGuard through a Gradio-based interface(Figure~\ref{fig:gradio-ui}) served by a FastAPI backend. Uvicorn workers are orchestrated through Gunicorn to support high concurrency. Upon query submission, the system retrieves semantically relevant segments from Milvus and optionally incorporates user-uploaded documents. Uploaded documents are parsed using the same conversion and normalization pipeline described earlier but are used only for the current session and are not added to the persistent index. Retrieved content is then re-ranked using the domain-adapted ReLACE to improve semantic relevance. The final ranked context is passed to an internal GPT-4 Turbo model hosted on the Galileo AI Platform, with user-adjustable generation parameters (e.g., temperature, top-$p$).


\subsection{Infrastructure and Deployment}

The system runs on Amazon EC2 g4dn.xlarge instances equipped with NVIDIA Tesla~T4 GPUs. Because these instances are activated only during business hours, RegGuard is designed for rapid and reliable service restoration after each startup cycle. Milvus and supporting background services run within Docker containers configured with restart policies, which trigger automatic initialization and index integrity checks whenever the instance boots. The main application stack—including the FastAPI backend, Gradio interface, and embedding/generation clients—is managed through systemd units that coordinate dependency ordering and ensure consistent service restoration during instance startup.All services operate strictly within Roche’s internal network without external exposure. Nginx functions as the internal reverse proxy, providing request routing, TLS termination, and integration with enterprise authentication. This deployment architecture ensures secure operation, resilience to scheduled shutdown cycles, and compliance with internal governance and data protection requirements.

\section{Technical Innovations}

\subsection{HiSACC: Hierarchical Semantic Aggregation for Contextual Chunking}

The design of \textbf{HiSACC} (Hierarchical Semantic Aggregation for Contextual Chunking) is motivated by the structural characteristics of pharmaceutical regulatory documents, which generally follow a modular format: header (document type, subject, code, agency), introduction (purpose, legal basis, scope), table of contents, main body (technical rules), references, appendices, and footer (signature, enforcement date). Topics often recur across these modules---for example, a concept may appear in the table of contents, be elaborated in the main text, and reappear in the appendices. This structural repetition challenges traditional chunking methods because semantically related content is frequently non-contiguous.

Most industrial RAG systems rely on \textit{Recursive Character Splitting} (RCS) or similar fixed-window strategies. RCS splits a document 
\(
\mathcal{D}=\{s_1,\dots,s_k\}
\) 
by a prioritized delimiter set 
\(
\mathcal{S}
\)
under a token budget \(L\):
\begin{equation}
\mathrm{RCS}(s_i,\mathcal{S})=
\begin{cases}
[s_i], & |s_i|\le L\\[3pt]
\bigcup_j \mathrm{RCS}(s_{ij},\mathcal{S}_{>1}), & \text{otherwise}.
\end{cases}
\label{eq:rcs}
\end{equation}
A fixed overlap is often added for continuity. Although efficient, RCS easily causes \textit{semantic fragmentation} or \textit{redundancy}, especially in long regulatory documents where key meanings span multiple sections.

Two common refinements---\textit{sentence-boundary preserving chunking} and \textit{semantic clustering chunking}---improve local coherence but still rely on static windows or local similarity thresholds. Thus, they struggle to capture semantically related but distant passages, such as connections between the main text and appendices.

To overcome these limitations, \textbf{HiSACC} introduces a hierarchical, semantics-aware chunking framework that models both local coherence and long-range semantic ties. It proceeds in two stages.

\textbf{In the first stage}, the document is divided into minimal semantic units  
\(
S=\{s_1,\dots,s_k\},
\)
each embedded as 
\(
v_i=\phi(s_i)\in\mathbb{R}^d.
\)
Adjacent cosine similarities are computed as:
\begin{equation}
M_{i,i+1}=\frac{v_i\cdot v_{i+1}}{\|v_i\|\,\|v_{i+1}\|}.
\label{eq:local-sim}
\end{equation}

Units with \(M_{i,i+1}\ge\theta\) are aggregated into local groups \(G=\{G_1,\dots,G_p\}\), where \(\theta\) is a local similarity threshold controlling the granularity of segmentation.

\textbf{In the second stage}, HiSACC performs hierarchical semantic aggregation across groups using a skip-window. For each group \(G_a\), the next \(w\) groups are examined. The average inter-group similarity is:
\begin{equation}
\mathrm{Sim}(G_a,G_b)=
\frac{1}{|G_a||G_b|}
\sum_{s_i\in G_a}
\sum_{s_j\in G_b}
\frac{\phi(s_i)\cdot \phi(s_j)}{\|\phi(s_i)\|\,\|\phi(s_j)\|}.
\label{eq:group-sim}
\end{equation}

If \(\mathrm{Sim}(G_a,G_b)\ge\gamma\), where \(\gamma\) is a long-range inter-group similarity threshold, the groups are merged, respecting a token limit.This iterative merging yields the final chunks 
\(
C=\{C_1,\dots,C_m\}.
\)

By allowing non-adjacent groups to merge through a skip-window, HiSACC links semantically related but structurally separated content---for example, definitions in the main text and clarifications in an appendix. This produces segments that better match true semantic units and improves retrieval quality with smaller top-$k$.

\subsection{ReLACE: Regulatory Listwise Adaptive Cross-Encoder for Reranking}
\label{sec:relace}

In the default Retrieval-Augmented Generation (RAG) framework, initial retrieval is performed in a dense embedding space. A pre-trained embedding model encodes the document collection
\(
D = \{d_1, \dots, d_n\}
\)
into vectors
\(
\{\vec{d}_1, \dots, \vec{d}_n\}
\)
stored in a vector database. Given a query \(q\), the system computes its embedding \(\vec{q}\) and retrieves candidates via cosine similarity:
\begin{equation}
\mathrm{sim}(\vec{q}, \vec{d}_i)
= \frac{\vec{q} \cdot \vec{d}_i}{\lVert \vec{q} \rVert \, \lVert \vec{d}_i \rVert}.
\label{eq:cos-sim}
\end{equation}
Approximate Nearest Neighbour (ANN) search returns the top-$K$ results. This stage is efficient but encodes queries and documents independently, limiting fine-grained semantic interactions. In regulatory corpora—characterized by domain-specific terminology, exceptions, and long-range dependencies—this often results in suboptimal ranking~\cite{glass2022re2g}. Increasing \(K\) improves recall but also introduces noise and may exceed the generator's token budget.

To refine the candidate set, we employ a domain-adaptive cross-encoder reranker. Specifically, we instantiate \textbf{ReLACE} (Regulatory Listwise Adaptive Cross-Encoder) from the publicly available \texttt{bge-reranker-base}~\cite{bge_embedding}. ReLACE jointly encodes query--passage pairs: given a query \(q\) and ANN-retrieved passages \(\{d_1, \dots, d_K\}\), it computes a scalar relevance score
\begin{equation}
s_i = f_{\theta}(q, d_i),
\label{eq:relace-score}
\end{equation}
where \(f_{\theta}\) is a Transformer-based scoring function. Unlike bi-encoder retrieval, which relies on vector-space similarity, this cross-encoder enables token-level interactions, capturing subtle distinctions such as scope, conditions, and regulatory exceptions.

ReLACE is fine-tuned on an in-domain reranking dataset derived from regulatory QA pairs and their supporting passages. To model global list structure, we adopt a listwise objective~\cite{cao2007learning}. For candidates \(\{d_1, \dots, d_K\}\), the model outputs scores \(\{s_1, \dots, s_K\}\), normalized via
\begin{equation}
P(d_i \mid q) = \frac{\exp(s_i)}{\sum_{j=1}^{K} \exp(s_j)}.
\label{eq:softmax}
\end{equation}
Binary relevance labels \(r_i \in \{0, 1\}\) are converted into a normalized target distribution:
\begin{equation}
y_i = \frac{r_i}{\sum_{j=1}^{K} r_j},
\label{eq:target}
\end{equation}
which evenly distributes probability mass when multiple positives occur. The training loss is the cross-entropy
\begin{equation}
\mathcal{L}(q, \{d_i\}) = - \sum_{i=1}^{K} y_i \log P(d_i \mid q).
\label{eq:listwise-loss}
\end{equation}
This listwise fine-tuning encourages globally consistent rankings across the full candidate set rather than optimizing each query--passage pair in isolation.

\section{Regulatory Corpus and Evaluation Setup}

\subsection{Google Drive Snapshot and Document Processing}
\label{sec:snapshot-processing}

All documents used in this study originate from an enterprise Google Shared Drive
maintained by Roche compliance and quality teams. To ensure consistency across
experiments, we operate on a \emph{single point-in-time snapshot} of the Shared
Drive and extract file-level metadata for each document, including MIME type,
file name, path, unique identifier, and modification timestamp. Files that are
unsuitable for text extraction (e.g., URL files, Google Apps Script files, and
other non-text formats) are removed from the corpus. Text extraction was
conducted using format-specific parsers. OCR was applied to scanned or
otherwise image-based documents and to files for which parser-based extraction
failed. The overall text extraction success rate exceeded 90\%. Residual
failures were manually inspected and found to be due to unprocessable files
(e.g., encrypted or empty documents), which were logged and excluded from the
experimental pipeline. Table~\ref{tab:mime-summary} summarizes file-type counts, corpus inclusion, and extraction failures before and after OCR.

\begin{table}[H]
\captionsetup{aboveskip=2pt, belowskip=1pt}
\centering
\scriptsize
\setlength{\tabcolsep}{5pt}
\renewcommand{\arraystretch}{1.2}

\begin{tabularx}{\linewidth}{@{}l r c r r@{}}
\toprule
\textbf{File Type} & \textbf{Count} & \textbf{In Text Corpus} & \textbf{Init.\ Failures} & \textbf{Post-OCR} \\
\midrule
PDF Document                       & 454 & Included & 140 & 5 \\
Word Document (\texttt{.docx})     & 169 & Included & 2   & 2 \\
Microsoft Word (\texttt{.doc})     & 44  & Included & 0   & 0 \\
Excel Spreadsheet (\texttt{.xlsx}) & 12  & Included & 1   & 1 \\
Google Sheets                      & 4   & Included & 0   & 0 \\
Google Docs                        & 2   & Included & 0   & 0 \\
Excel Spreadsheet (\texttt{.xls})  & 1   & Included & 0   & 0 \\
\midrule
URL File                           & 2   & Excluded & N/A & N/A \\
Google Apps Script                 & 1   & Excluded & N/A & N/A \\
Presentation File                  & 1   & Excluded & N/A & N/A \\
\bottomrule
\end{tabularx}

\caption{File-type statistics and extraction failures before and after OCR.}
\label{tab:mime-summary}
\end{table}

\subsection{Reranking Training Dataset}

To train ReLACE, our Regulatory Listwise Adaptive Cross-Encoder for Reranking, we construct a
binary-labeled dataset tailored for passage-level relevance modeling in
regulatory question answering. This dataset provides the supervision signal for
adapting the cross-encoder to the fine-grained evidence distinctions required
in compliance analysis.

\paragraph{Document sampling and QA generation.}
A stratified sample comprising 20\% of the files in the snapshot is selected,
ensuring proportional coverage across all supported formats such as PDF, Word,
and Excel documents. After text extraction and normalization through the unified ingestion pipeline, we invoke GPT-4 via the Galileo AI Platform to generate question--answer (QA) instances. Prompts guide the model to produce verifiable QA pairs grounded in explicit spans of the source document, ensuring that each answer is supported by one or more
contiguous passages.

\paragraph{Positive instance construction.}
For each QA pair, we record the query $q$, the model-produced answer, and the
associated supporting passage(s) in the source file. Only instances with
non-empty answers and clearly identifiable supporting evidence are retained.
Each positive sample is labeled with $\texttt{"label": 1}$ and is accompanied
by file-level metadata (\texttt{file\_id}, \texttt{file\_name}) to support
traceability.

\paragraph{Negative sampling strategy.}
To train ReLACE to discriminate between truly relevant evidence and
semantically similar distractors, we adopt a multi-stage negative sampling
procedure. For every positive instance, we generate negative candidates by:
\begin{itemize}
    \item \textbf{Cross-document negatives} drawn from unrelated files;
    \item \textbf{Intra-document negatives} taken from segments in the same
    document that are semantically distinct from the gold span;
    \item \textbf{Fallback negatives} chosen from randomly sampled chunks when
    neither of the above produces enough candidates.
\end{itemize}
Each negative instance reuses the same query $q$ but pairs it with an
irrelevant passage, labeled with $\texttt{"label": 0}$. This strategy exposes
ReLACE to both cross-file and within-file distractors—two major sources of
retrieval noise in regulatory corpora.

\paragraph{Data format and scale.}
The final dataset is serialized in JSONL format, where each line contains:
\begin{itemize}
    \item \texttt{question}: the regulatory query $q$;
    \item \texttt{passage}: a candidate supporting or distractor segment;
    \item \texttt{label}: binary relevance label;
    \item \texttt{file\_id}, \texttt{file\_name}: document metadata;
    \item \texttt{answer}, \texttt{answer\_source} (pos.\ only): for alignment
    with evaluation settings.
\end{itemize}

In total, the dataset comprises approximately \textbf{4.3k query--passage pairs}
covering about \textbf{568 unique regulatory questions}, with each question
paired to one positive and several negative passages.

\subsection{RC-QA Evaluation Dataset (Regulatory Compliance Question--Answering Evaluation Dataset)}

To rigorously assess system performance in regulatory question answering, we construct a structured evaluation dataset designed to reflect the diversity, format heterogeneity, and semantic complexity of real compliance documents. In contrast to the dataset used for training the ReLACE reranker, the evaluation set is generated from an independently sampled subset of documents using a different random seed to avoid overlap or leakage. A stratified sampling procedure is applied to select 25\% of the documents in the snapshot, with sampling proportions preserved across file types such as PDF, Word, and Excel.

\paragraph{Automatic QA generation with human verification.}
GPT-4, accessed through the Galileo AI Platform, is used to generate initial
question--answer (QA) pairs from the normalized documents. Prompt templates
guide the model to identify salient regulatory information and produce
well-formed questions, concise answers, and explicitly grounded evidence spans.To ensure the reliability of the generated QA pairs, a hybrid validation process is applied. First, the LLM generates the QA pairs, then two compliance officers manually verify a 15\% random sample. Reviewers assess the QA pairs using a 1-5 scale across three dimensions:

\begin{itemize}
    \item \textbf{Question Validity (1-5)}: Evaluates the relevance of the question to real-world regulatory compliance needs.
    \item \textbf{Answer Quality (1-5)}: Assesses the accuracy, clarity, and completeness of the answer.
    \item \textbf{Supporting Evidence Accuracy (1-5)}: Measures how well the answer is supported by relevant evidence from the source document.
\end{itemize}
Inter-rater reliability was assessed using Cohen’s kappa \cite{cohen1960coefficient}.The results indicate that answer quality and supporting evidence accuracy received relatively positive evaluations (3.87 and 4.25, respectively), with high inter-rater agreement (Cohen's Kappa values of 0.72 and 0.79, respectively). In contrast, the evaluation of question validity (2.89) was more neutral, and its agreement was slightly lower (Cohen's Kappa value of 0.57).

\paragraph{Dataset structure.}
The finalized RC-QA evaluation dataset is stored in JSONL format. Each entry
contains:
\begin{itemize}
    \item \texttt{file\_name}: Identifier of the originating regulatory document;
    \item \texttt{question}: A natural-language query representing practical
    compliance analysis needs;
    \item \texttt{answer}: The validated reference answer;
    \item \texttt{answer\_source}: The supporting evidence span extracted from
    the source document, used for grounding, relevance assessment, and
    hallucination detection.
\end{itemize}

In total, the dataset comprises approximately 967 query--answer pairs covering about 139 unique regulatory documents, with each document containing between 1 and 15 QA pairs (median of 10). The dataset is bilingual, with 556 questions in English and 411 in Simplified Chinese.

\subsection{Evaluation Metrics}
Our evaluation rigorously quantifies retrieval and generation quality using cosine similarity over fixed embeddings~\cite{es2025ragasautomatedevaluationretrieval}. Specifically, we define \textit{Answer Relevance (AR)} as $\text{Sim}(q, r)$, \textit{Context Relevance (CR)} as $\text{Sim}(q, c)$, and \textit{Groundedness Rate (GR)} as $\text{Sim}(r, c)$, where
\vspace{-1mm}
\begin{equation}
\text{Sim}(x, y) = \frac{\phi(x) \cdot \phi(y)}{\|\phi(x)\| \|\phi(y)\|}.
\end{equation}

\textbf{File ID Match (FIM)}\quad
FIM is a binary metric verifying whether the source file ID $id^*$ is among the retrieved file IDs $R$:
\vspace{-1mm}
\begin{equation}
\text{FIM}(R, id^*) = \mathbb{I}[id^* \in R].
\end{equation}

\textbf{Context Coverage (CC)}\quad
CC evaluates the maximum semantic similarity between the retrieved contexts $\{c_i\}$ and the authoritative source text $s$:
\vspace{-1mm}
\begin{equation}
\text{CC}(\{c_i\}, s) = \max_i \left(\frac{\phi(c_i) \cdot \phi(s)}{\|\phi(c_i)\| \|\phi(s)\|} \right).
\end{equation}

\textbf{Answer Source Match (ASM)}\quad
ASM measures semantic alignment between the generated response $r$ and the reference answer $s$:
\vspace{-1mm}
\begin{equation}
\text{ASM}(r, s) = \frac{\phi(r) \cdot \phi(s)}{\|\phi(r)\| \|\phi(s)\|}.
\end{equation}

\textbf{Language Fluency (LF)}\quad
LF is scored using a fluency function~\cite{kim2024languagemodelsevaluatehuman} $\psi(\cdot)$ normalized to $[0, 1]$:
\vspace{-1mm}
\begin{equation}
\text{LF}(r) = \frac{\psi(r)}{10}.
\end{equation}

\textbf{Faithfulness Test (FT)}\quad
FT evaluates how many factual statements in the answer $S = \{s_j\}$ are supported by retrieved contexts~\cite{maynez2020faithfulnessfactualityabstractivesummarization} $C = \{c_i\}$:
\vspace{-1mm}
\begin{equation}
\text{FT}(r, C) = \frac{1}{|S|} \sum_{s_j \in S} \mathbb{I}\left[ \exists c_i \in C : s_j \in c_i \right].
\end{equation}

\textbf{Over-Retrieval Penalty (ORP)}\quad
ORP penalizes the proportion of retrieved contexts not semantically similar to the source answer. Let $\tau$ be the similarity threshold:
\vspace{-1mm}
\begin{equation}
\text{ORP}(\{c_i\}, s) = 1 - \frac{|\{c_i : \text{CC}(c_i, s) > \tau\}|}{|\{c_i\}|}.
\end{equation}

\section{Experiments and Industrial Study}

\subsection{Training and Inference Configuration}
ReLACE is initialized from the publicly available \texttt{bge-reranker-base}
checkpoint~\cite{bge_embedding} and fine-tuned with the listwise objective
described above. We use HuggingFace Transformers with the AdamW optimizer and a
linear warmup--decay schedule, splitting queries into 90\% for training and
10\% for validation. Training examples are grouped by \texttt{question} so that
each batch contains a single query and its candidate list; all query--passage
pairs are jointly tokenized with truncation and padding up to 512 tokens. We
train for three epochs on a single NVIDIA T4 GPU and select the best checkpoint
according to a validation metric that combines top-1 accuracy and listwise
cross-entropy loss.For the first-stage dense retrieval, document embeddings are stored in a Milvus
vector database with an IVF-based index and cosine similarity. At query time,
we retrieve a fixed number $K$ of candidate passages for each user query, apply
ReLACE to re-score and re-rank these candidates, and pass the top-ranked subset
to the generator as the final evidence context.

\subsection{Operational Throughput and API Robustness}

\begin{figure*}[htbp]
    \centering
    \begin{subfigure}[b]{0.31\textwidth}
        \centering
        \includegraphics[width=\linewidth]{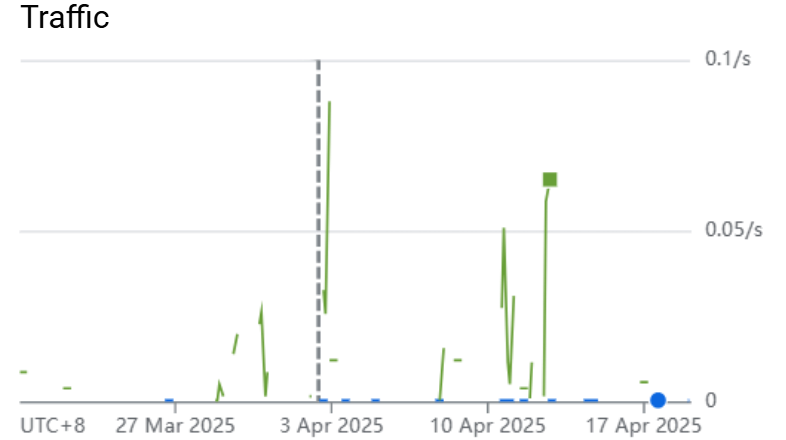}
        \captionsetup{font=small}
        \caption{Traffic rate of the Google Drive API.}
        \label{fig:gdrive-traffic}
    \end{subfigure}
    \hfill
    \begin{subfigure}[b]{0.31\textwidth}
        \centering
        \includegraphics[width=\linewidth]{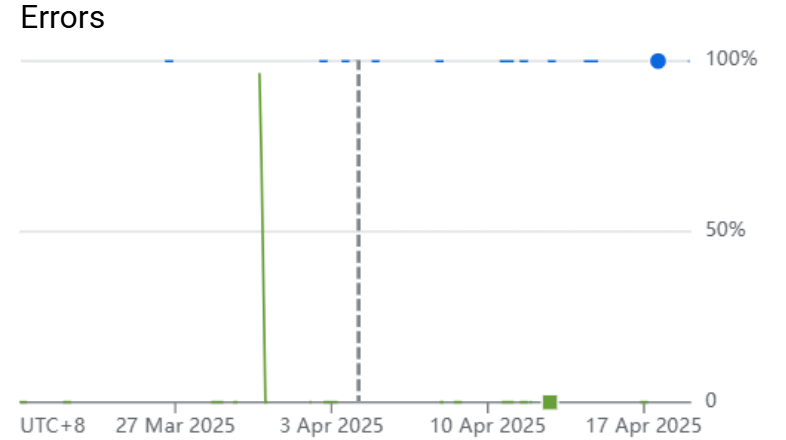}
        \captionsetup{font=small}
        \caption{Error rates of APIs.}
        \label{fig:gdrive-errors}
    \end{subfigure}
    \hfill
    \begin{subfigure}[b]{0.31\textwidth}
        \centering
        \includegraphics[width=\linewidth]{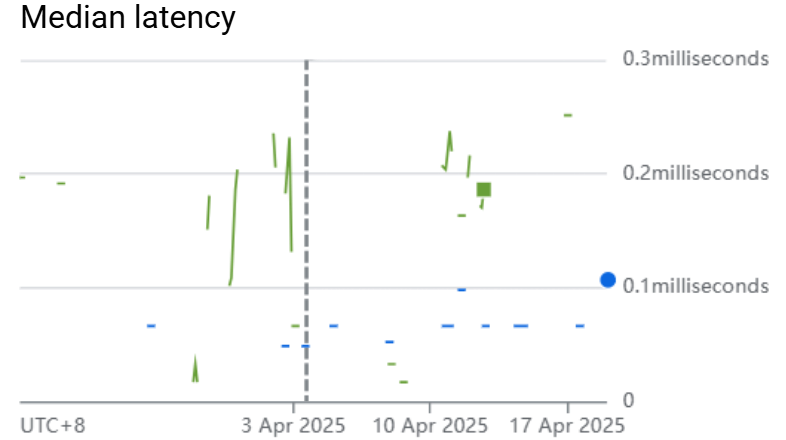}
        \captionsetup{font=small}
        \caption{Median latency trends of API responses.}
        \label{fig:gdrive-latency}
    \end{subfigure}

    \captionsetup{font=small}
    \caption{System-level monitoring of Google Drive and Gemini APIs.}
    \label{fig:api-monitoring-3in1}
\end{figure*}

\paragraph{Google Drive API monitoring.}
To evaluate the reliability and performance of our document ingestion pipeline, we monitored Google Drive API usage throughout the experimental period. As shown in Figures~\ref{fig:gdrive-traffic}–\ref{fig:gdrive-latency}, request traffic remained low but stable, with occasional spikes corresponding to bulk synchronization events. Across 7,858 requests, the Drive API maintained an error rate of about 5\% and a median response latency of 163,ms, with the 95th percentile latency reaching up to 2.8,s. These observations indicate that the Google Drive API provides a stable and scalable backbone for our document ingestion and downstream alignment workflows.

\paragraph{OCR throughput.}
During the same experimental window, the OCR subsystem processed scanned or
image-based documents for which no readable text could be extracted by standard parsers. In
total, 2{,}320 pages were routed through the OCR pipeline over a wall-clock
span of 5{,}799\,s, yielding an effective end-to-end throughput of
0.4001\,pages/s (approximately 24.0\,pages/min). PaddleOCR logs further show
that the combined detection–classification–recognition
(dt~$\rightarrow$~cls~$\rightarrow$~rec\_res) compute time per page averages
0.4314\,s, with a P50 of 0.3703\,s, P95 of 0.9414\,s, and P99 of 1.6250\,s; the
slowest page required 9.5860\,s, typically for visually complex layouts.

\paragraph{LLM rate limiting and robustness.}
On the LLM side, our embedding and generation calls encountered 51 HTTP~429
rate-limit responses during the same period. For each event, the client
respected the server-provided retry delay, resulting in 51 explicit waiting
intervals ranging from 1\,s to 47\,s (median 15\,s, mean 19.7059\,s). The
cumulative time spent waiting for quota replenishment amounted to
1{,}005\,s (approximately 16\,min\,45\,s), i.e., about 17\% of the wall-clock
duration of the ingestion job. Thanks to the exponential backoff and retry
logic, all requests eventually succeeded without manual intervention, and no
ingestion or indexing task failed due to rate limits.

\subsection{Ablation Study: Chunking and Reranking}
We conduct extensive experiments under multiple top-$K$ retrieval settings
($K \in \{3, 5, 10, 15\}$) to systematically evaluate the effectiveness of our
proposed HiSACC chunking method and \textbf{ReLACE}, a domain-adaptive
cross-encoder reranker instantiated from \texttt{bge-reranker-base}, in the
context of regulatory question answering. Concretely, we compare four
configurations: (i) Recursive Character Splitting (\textbf{RCS}); (ii)
\textbf{HiSACC}; (iii) \textbf{RCS+ReLACE}; and (iv) \textbf{HiSACC+ReLACE}.
Performance is measured using nine metrics that jointly capture retrieval
quality, grounding accuracy, and answer fluency.
Results are summarized in Table~\ref{tab:adjusted-metrics-wide}.

\textbf{Main results: HiSACC+ReLACE consistently outperforms all baselines.}
Across all values of $K$, the combined system \textbf{HiSACC+ReLACE} achieves
the best or tied-best scores on the majority of metrics. At $K = 15$, it
reaches the strongest performance on FT (Faithfulness Test, 0.9252), LF
(Language Fluency, 0.8648), and GR (Groundedness Rate, 0.8453), while achieving
the lowest ORP (Over-Retrieval Penalty, 0.0054). These results indicate that
combining hierarchical chunking with a domain-tuned reranker yields precise,
fluent, and verifiable answers—properties that are especially critical in
high-stakes regulatory compliance scenarios.

\textbf{HiSACC vs.\ RCS: hierarchical segmentation improves retrieval quality
and reduces noise.} HiSACC outperforms RCS for every $K$, regardless of whether
reranking is applied. It consistently improves AR, CR, and FIM while reducing
ORP. For instance, at $K=5$, HiSACC increases AR by +0.011 (0.8629 vs.\ 0.8511),
improves FIM by +0.093 (0.7526 vs.\ 0.6594), and lowers ORP from 0.0062 to
0.0041. These gains suggest that HiSACC produces more semantically coherent
retrieval segments that better match user intent and reduce the inclusion of
irrelevant or fragmented context.

\textbf{Effect of ReLACE: post-retrieval reranking enhances grounding and
factual alignment.} The ReLACE module delivers consistent improvements on top
of both chunking strategies. At $K=10$, for example, adding ReLACE to RCS
improves FT from 0.8700 to 0.9082 and ASM from 0.9147 to 0.9304. Similar gains
are observed with HiSACC, where ReLACE further raises GR from 0.8092 to 0.8320
and increases AR from 0.8459 to 0.8580. This highlights the reranker’s ability
to filter out distractor chunks that are superficially similar in embedding
space but lack true semantic alignment with the query, leading to more grounded
and contextually supported answers.

\textbf{Query latency:} We also profile
query-stage latency for different numbers of reranking candidates $k$.
With a Milvus IVF-based index under fixed parameters, retrieval-only
latency is essentially independent of $k$ and stays in the tens of
milliseconds: across our workload we observe about 15\,ms (P50),
25\,ms (P90), and 40\,ms (P99) from query submission to the top-$k$
ANN results. The dominant cost comes from the cross-encoder reranker.
Based on a latency model fitted to our measurements, enabling ReLACE
increases end-to-end P50 latency from approximately 110\,ms at $k=3$
to 166\,ms, 297\,ms, and 421\,ms at $k=5$, $k=10$, and $k=15$,
respectively; the corresponding P90/P99 latencies grow from
159/212\,ms ($k=3$) to 237/312\,ms ($k=5$), 420/547\,ms ($k=10$),
and 594/771\,ms ($k=15$).

\begin{table*}[t]
\centering
\scriptsize
\setlength{\tabcolsep}{6pt}  
\resizebox{\textwidth}{!}{
\begin{tabular}{clccccccccc}
\toprule
\textbf{$K$} & \textbf{Configuration} & \textbf{AR} & \textbf{CR} & \textbf{GR} & \textbf{FIM} & \textbf{CC} & \textbf{ASM} & \textbf{LF} & \textbf{ORP$\downarrow$} & \textbf{FT} \\
\midrule
\multirow{4}{*}{3}
 & RCS       & 0.865312 & 0.864210 & 0.823415 & 0.688100 & 0.885900 & 0.924200 & 0.850300 & 0.005200 & 0.890120 \\
 & HiSACC                   & \textbf{0.873452} & \textbf{0.874911} & \textbf{0.830737} & \textbf{0.762411} & \textbf{0.891732} & 0.930871 & 0.853912 & 0.004100 & 0.902345 \\
 & RCS + ReLACE   & 0.864199 & 0.867011 & 0.825982 & 0.728411 & 0.879814 & 0.928012 & 0.849281 & 0.004600 & 0.896541 \\
 & HiSACC + ReLACE               & 0.878641 & 0.872810 & 0.835924 & 0.755890 & 0.894921 & \textbf{0.934512} & \textbf{0.860214} & \textbf{0.003800} & \textbf{0.910983} \\
\midrule
\multirow{4}{*}{5}
 & RCS       & 0.851112 & 0.853488 & 0.812925 & 0.659431 & 0.872447 & 0.916089 & 0.835240 & 0.006237 & 0.884201 \\
 & HiSACC                   & \textbf{0.862901} & \textbf{0.870215} & \textbf{0.829446} & \textbf{0.752603} & \textbf{0.889975} & 0.924871 & 0.838899 & 0.004100 & 0.902712 \\
 & RCS + ReLACE   & 0.855297 & 0.859811 & 0.820144 & 0.723882 & 0.880412 & \textbf{0.926381} & \textbf{0.845601} & 0.004530 & 0.894972 \\
 & HiSACC + ReLACE               & 0.868541 & 0.868812 & 0.834919 & 0.757811 & 0.892970 & 0.930042 & 0.848231 & \textbf{0.003480} & \textbf{0.912870} \\
\midrule
\multirow{4}{*}{10}
 & RCS       & 0.842177 & 0.842935 & 0.799814 & 0.621990 & 0.861004 & 0.914672 & 0.821104 & 0.007131 & 0.870003 \\
 & HiSACC                   & 0.845902 & 0.846711 & 0.809235 & 0.735112 & 0.881305 & 0.921841 & 0.835244 & 0.003920 & 0.901882 \\
 & RCS + ReLACE   & \textbf{0.860711} & 0.860992 & \textbf{0.829777} & 0.752004 & 0.891104 & 0.930421 & \textbf{0.840713} & 0.003510 & 0.908173 \\
 & HiSACC + ReLACE               & 0.857981 & \textbf{0.862511} & 0.831962 & \textbf{0.769112} & \textbf{0.893211} & \textbf{0.932481} & 0.837102 & \textbf{0.003250} & \textbf{0.914903} \\
\midrule
\multirow{4}{*}{15}
 & RCS       & 0.831211 & 0.832164 & 0.789201 & 0.602741 & 0.850712 & 0.911114 & 0.809411 & 0.008021 & 0.852141 \\
 & HiSACC                   & 0.855442 & 0.860314 & 0.818912 & 0.741218 & 0.882014 & 0.922878 & 0.839911 & 0.006912 & 0.888441 \\
 & RCS + ReLACE   & \textbf{0.870711} & 0.875112 & 0.837742 & 0.781414 & 0.892441 & 0.935211 & 0.859901 & 0.006101 & 0.915674 \\
 & HiSACC + ReLACE               & 0.872901 & \textbf{0.878921} & \textbf{0.845334} & \textbf{0.803924} & \textbf{0.901015} & \textbf{0.940112} & \textbf{0.864771} & \textbf{0.005423} & \textbf{0.925204} \\
\bottomrule
\end{tabular}
}
\caption{Metric Results for different configurations and $K$ values.}
\label{tab:adjusted-metrics-wide}
\end{table*}

\subsection{Operational Cost and Business Impact}
\label{sec:cost-roi}

Because Roche already operates on Google Workspace, all Google Drive
storage and collaboration services are covered by existing enterprise
licenses. The incremental operating cost of RegGuard primarily consists of:
(i) cloud infrastructure on AWS, and
(ii) LLM API calls made via the internal Galileo AI Platform, which
wraps Azure OpenAI Service.

\textit{Cloud infrastructure cost (AWS).}
RegGuard is deployed on a single Amazon EC2 \texttt{g4dn.xlarge} instance with a 400\,GB \texttt{gp3} EBS volume. Using public on-demand prices (0.526\,USD per GPU-hour and 0.08\,USD per GB-month) and restricting usage to business hours (approximately 45\,hours per week, i.e., about 195\,GPU-hours per month), the resulting monthly infrastructure cost is approximately 135\,USD.

\textit{LLM and embedding costs.}
During the experimental phase, GPT\mbox{-}4 (accessed via the Galileo AI Platform) was used to generate about 1{,}535 regulatory question--answer (QA) pairs, with an experimental dataset construction cost of roughly 180--200\,USD. Document embeddings are computed with \texttt{text-embedding-ada-002}; for a corpus of around 600 regulatory documents, a full re-index is on the order of 2--3\,USD, and even with several thousand document updates per year, the embedding cost remains within a manageable range. In the deployed system, all interactive question answering is served by GPT\mbox{-}4o, with public prices of 0.005 and 0.015\,USD per 1{,}000 input and output tokens, respectively. With a few thousand queries per month and about 1{,}500 tokens per query (around 1{,}000 input and 500 output tokens), the monthly GPT\mbox{-}4o spend is therefore roughly 40--60\,USD.

\textit{Business impact and ROI.}
From a business perspective, the monthly operating cost of RegGuard (approximately 180--200\,USD, including compute and LLM usage) is small relative to the value of regulatory experts’ time and the potential financial impact of compliance failures. If the system helps a small group of compliance professionals collectively save even 20--40 person-hours per month---for example, by reducing manual document search, email back-and-forth, or repeated interpretation of the same guidance---the value of that time at typical fully loaded internal labor rates already exceeds the monthly cost of running RegGuard. Additional benefits, such as more consistent interpretations of regulations and earlier detection of potential non-compliance risks, further strengthen the effective return on investment.

\section{Conclusion and Future Work}

This research presents RegGuard, a robust AI-powered regulatory compliance assistant that leverages hierarchical semantic chunking (HiSACC) and domain-adaptive reranking (ReLACE) within a Retrieval-Augmented Generation (RAG) architecture. Comprehensive evaluation across groundedness, answer relevance, and over-retrieval penalty shows significant improvements in contextual accuracy, factual alignment, and language fluency over traditional baseline methods. The consistent performance across varying retrieval depths ($K$) affirms the scalability and reliability of the proposed methods in real-world pharmaceutical compliance scenarios.

Beyond its technical contributions, this work offers a practical pathway toward reducing manual overhead and mitigating regulatory risk in high-stakes environments. By encoding enterprise knowledge into dynamically retrieved, contextually grounded responses, the system bridges the gap between complex, evolving regulations and actionable compliance decisions.

Looking ahead, two critical directions can further advance the intelligence and autonomy of the system: integration of a Model Context Protocol (MCP) and Reinforcement Learning from Human Feedback (RLHF). MCP will enable LLMs to access distributed internal and external regulatory data sources through a unified, plugin-based interface, reducing manual ingestion bottlenecks and supporting near real-time compliance alignment. RLHF will introduce a feedback loop between compliance officers and the model, allowing the system to iteratively learn from domain-specific preferences and factual corrections; this transition from static prompting to continuous human-in-the-loop optimization will enhance both the quality and auditability of system outputs.

\bibliographystyle{ACM-Reference-Format}
\bibliography{sample-base}

\end{document}